\documentclass[letterpaper]{article} 
\usepackage[submission]{aaai23}  
\usepackage{times}  
\usepackage{helvet}  
\usepackage{courier}  
\usepackage[hyphens]{url}  
\usepackage{graphicx} 
\usepackage{natbib}  
\usepackage{caption} 
\usepackage{algorithm}
\usepackage{algorithmic}

\usepackage{newfloat}
\usepackage{listings}
\DeclareCaptionStyle{ruled}{labelfont=normalfont,labelsep=colon,strut=off} 
\floatstyle{ruled}
\newfloat{listing}{tb}{lst}{}
\floatname{listing}{Listing}
 \usepackage{hyperref} 

\title{Skeleton-based Action Recognition via Temporal-Channel Aggregation}
\author{
    Shengqin Wang\textsuperscript{\rm 1},
    Yongji Zhang\textsuperscript{\rm 2},
    Minghao Zhao\textsuperscript{\rm 3},
    Hong Qi\textsuperscript{\rm 4},
    Kai Wang\textsuperscript{\rm 5},
    Fenglin Wei\textsuperscript{\rm 6},
    Yu Jiang \thanks{*Corresponding author}
}

\affiliations{

}

\usepackage{bibentry}
\usepackage{amsmath,amsfonts}
\usepackage{algorithmic}
\usepackage{algorithm}
\usepackage{array}
\usepackage[caption=false,font=normalsize,labelfont=sf,textfont=sf]{subfig}
\usepackage{amsmath}
\usepackage{textcomp}
\usepackage{stfloats}
\usepackage{url}
\usepackage{multirow}
\usepackage{verbatim}
\usepackage{graphicx}

\usepackage{lineno}

\usepackage{cite}
\usepackage{float}
\usepackage{color}
\definecolor{BlueColor}{rgb}{0.0, 0.0, 1.0}
\definecolor{RedColor}{rgb}{1.0, 0.0, 0.0}

\begin{document}
\maketitle
\begin{abstract}
Skeleton-based action recognition methods are limited by the semantic extraction of spatio-temporal skeletal maps. However, current methods have difficulty in effectively combining features from both temporal and spatial graph dimensions and tend to be thick on one side and thin on the other. In this paper, we propose a Temporal-Channel Aggregation Graph Convolutional Networks (TCA-GCN) to learn spatial and temporal topologies dynamically and efficiently aggregate topological features in different temporal and channel dimensions for skeleton-based action recognition. We use the Temporal Aggregation module to learn temporal dimensional features and the Channel Aggregation module to efficiently combine spatial dynamic channel-wise topological features with temporal dynamic topological features. In addition, we extract multi-scale skeletal features on temporal modeling and fuse them with an attention mechanism. Extensive experiments show that our model results outperform state-of-the-art methods on the NTU RGB+D, NTU RGB+D 120, and NW-UCLA datasets.
\end{abstract}

\section{Introduction}



Action recognition is an essential component of computer vision and a very active research topic. The development of sensors~\cite{2016NTU, 2020NTU} and advanced algorithms for human pose estimation~\cite{2017RMPE, CaoZhe0OpenPose} has made obtaining accurate 3D skeletal data easier, especially when the skeletal data are relatively computationally small and robust against complex backgrounds and changing conditions such as body size, viewpoint, and motion speed~\cite{DBLP:conf/aaai/YanXL18, 2020sttr, 20133D, 2017Going}. Skeleton-based human action recognition is thus of great interest.
\begin{figure}[t] 
\centering 
\includegraphics[width=1\linewidth]{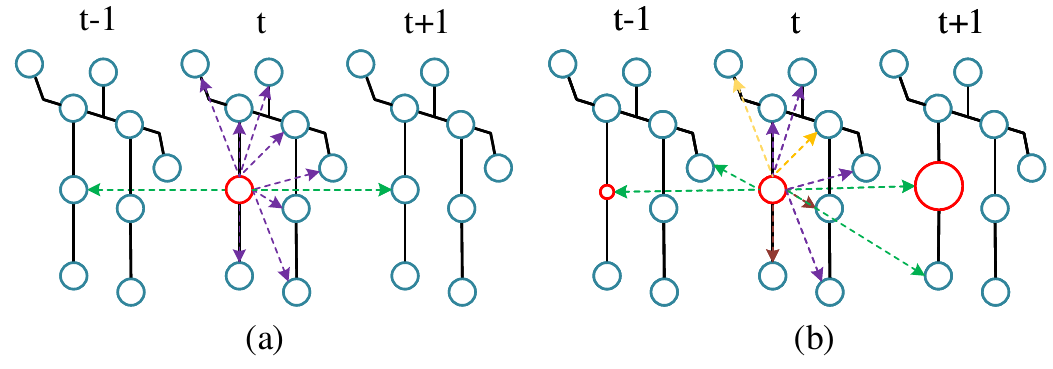} 
\caption{Processing spatio-temporal skeletal features with TCA-GCN. A simplified analysis is performed with the red joints as target nodes. Figure (a) shows the aggregation operation in space through a priori topological relations (purple arrows). In time the node uses fixed (green arrows) node characteristics. Figure (b) shows our TCA module aggregation in space using a combination of dynamic and static topology (different colored arrows). In time  node uses features that are dynamic (different node sizes), which are obtained by aggregating weights on the time dimension.} 
\label{fig11} 
\end{figure}
The joints of the skeletal spatio-temporal graph of human action are correlated. Yan et al.~\cite{DBLP:conf/aaai/YanXL18} applied graph convolutional networks (GCNs) to the field of skeleton-based action recognition for the first time. They used GCNs and temporal convolution to extract motion features and graph structures to model the correlation between human joints. For example, the idea of learnable skeletal edge weights was proposed, and the problem of different importance of different joints in action recognition was initially explored. Based on this, some approaches~\cite{2018Two, 2020Skeletonshift, qin2021fusing} exploit second-order information of skeletal data (length and orientation of bones, angles) and use data augmentation in the form of multi-stream networks. However, these approaches make it difficult to model the relationship between unnaturally connected joints, limiting the representational power of the model. Some recent approaches~\cite{2018Two, 2020sttr, 2019Semantics,peng2020learning} use attentional or other mechanisms to model the human skeletal structure adaptively, however, the variability of the different channel topologies is not taken into account.
Yu et al.~\cite{2021Channel} proposed channel topology refinement graph convolution (CTR-GC) to learn topologies and aggregate features in different channel dimensions dynamically. However, these methods ignore the balance of spatial and temporal dimensions, and they mostly favor modeling in the spatial dimension. Although using an attention mechanism in the temporal dimension can effectively model the correlation of key points in the temporal dimension, the combination of spatial and temporal features is still simple.

 In the previously proposed graph convolutional neural network, it mainly consists of a series of operations on the filter parameter matrix, the input feature matrix, and the adjacency matrix, the diagonal matrix of the graph structure. For filter parameter matrix in the field of skeletal action recognition, there are various processing methods. In the area of spatial modeling, Spatial-Temporal Graph Convolutional Networks (ST-GCN)~\cite{2018Spatial} first applied the graph convolutional neural network to the field of skeletal action recognition, which proposed the idea of learnable skeletal edge weights to solve the problem of different importance of different parts due to the grouping change between joints during human movement, for example, during the upper body joint parts may play a more important role than the lower body joint parts during writing. For other models~\cite{2018Two,2019Spatial,2021Channel}, most of them adopt a simple multi-channel convolution method to achieve a high-dimensional representation to solve the problem of feature transformation of the input data. However, for this part of the processing of the actual method mostly makes the skeletal sequence matrix has the disadvantage of temporal invariance, for each frame of the skeleton is constantly changing dynamically, can not be more targeted for the adaptive weight calibration characteristics of the input feature matrix. For the temporal modeling aspect of the skeletal sequence, for the skeletal spatio-temporal map constructed by connecting joints between consecutive frames, many existing methods~\cite{2018Spatial,2019Spatio,2018Two,2020Skeleton,2019Actional} have used a fixed temporal kernel size method to process, with the multi-scale learning method proposed in Disentangling and Unifying Graph Convolutions (MS-G3D)~\cite{2020Disentangling} makes the model obtain a larger feeling field, realizes the modeling of different time sequences, and uses residual connections~\cite{2016Deep} to facilitate training. To reduce the inference speed, a multi-scale temporal modeling module with less analysis as well as residual connectivity was used in Channel-wise Topology Refinement Graph Convolution (CTR-GCN)~\cite{2021Channel}. However, for temporal scale learning, it is difficult to solve issues such as skeletal action semantics for better effective modeling by simply adding to achieve feature fusion. For the multi-stream fusion framework problem, most of the previous methods~\cite{2021Channel,DBLP:conf/cvpr/Cheng0HC0L20,2020Dynamic} of fixed weights, we found that there is a problem of inconsistent use of stream in the case of optimal use results, and for stream using the same weights, which hardly reflects the importance of different streams.

To address the above problem, we propose a method (temporal-channel adaptive aggregation) that can dynamically learn spatio-temporal features of the skeleton and effectively combine the two. Specifically, we use contextual information in the temporal dimension of the skeletal map to assist in calibrating the weights generated by the input adaptively and perform feature aggregation in the temporal dimension. This helps us establish skeletal associations across different time series and finally obtain a dynamic temporal feauture representation in high-dimensional space. On the other hand, we use channel-wise topology modeling to learn the dynamic spatial topology of the input information, which is then aggregated with the previously obtained dynamic temporal topology in the channel dimension. In addition, we use a temporal modeling module with a multi-scale feature fusion mechanism as a complement. It first performs multi-scale features extraction of the aggregated spatio-temporal topology to obtain a larger perceptual field, and then performs feature fusion with an attention mechanism. The temporal modeling module can effectively implement topological feature fusion to help us complete the modeling of the action. Its framework is shown in Figure~\ref{fig11}(b).

Overall, our main contributions are mainly in the following aspects:

\begin{itemize}
    \item  We propose a Temporal-Channel Aggregation Graph Convolutional Networks (TCA-GCN) to learn the topology in space and time dynamically and efficiently aggregate joint features in different temporal and channel dimensions for skeleton-based action recognition. Our approach can effectively balance features in the temporal and spatial dimensions of the human skeletal graph.
    \item We validated the validity of our model on three publicly available datasets, NTU RGB+D~\cite{2016NTU}, NTU RGB+D 120~\cite{2020NTU}, and NW-UCLA~\cite{DBLP:journals/corr/WangNXWZ14}, and performed extensive ablation experiments.
\end{itemize}

\section{Related Work}

{\bf Skeleton-based Action Recognition}. Most of the early skeleton-based motion recognition was done by hand-made feature-based methods~\cite{2014Human,2015Modeling}  to simulate human motion. However, these methods mainly used the idea of translation between joints and 3D rotation, which suffers from incomplete considerations and performance problems~\cite{2021Learning}. In recent years, with the development of deep learning, more and more neural network structures have been applied in skeleton-based motion recognition, among which the more widely used ones are classified as recurrent neural networks (RNN), convolutional neural networks (CNN), and graph convolutional networks (GCN).

RNN-based methods are mainly targeted at processing sequential data, and these methods extract features of bones and then construct skeletal sequence-related dependencies~\cite{2015Hierarchical,2017Adaptive,2016Spatio,2016NTU,2017View}. Shahrou et al.~\cite{2016NTU} proposed a generic action recognition dataset and also proposed a part-aware LSTM model to model the long-term temporal correlation of each body part feature. Zhang et al.~\cite{2017View} proposed an adaptive recurrent neural network based on LSTM structure based on action recognition related features that can adapt to the most appropriate observation view. The RNN-based approach is more difficult to train and less parallel. CNN-based methods for skeletal action recognition have higher efficiency. CNN-based methods~\cite{2017Two,2017Interpretable,2017A,Liu2017Enhanced,2017Skeleton,2017Skeleton1} usually convert the skeletal sequence coordinates into three channels and then classify the features extracted through the network. Xu et al.~\cite{xu2022topology} proposed a pure CNN structure that can improve the modeling of irregular skeleton topology. However, these methods are difficult to achieve full utilization of the natural topology of the bones, and GCN can make full use of the spatio-temporal information of the skeleton to model the topology more effectively.

The ST-GCN proposed by Yan et al.~\cite{DBLP:conf/aaai/YanXL18} introduced graph covolutional networks (GCNs) to the field of skeletal action recognition for the first time. The ST-GCN uses three joint segmentation methods by utilizing data information such as skeletal space-time. The method also proposes the Temporal Convolutional Network (TCN) for modeling the temporal dimension. Based on this approach a large number of methods based on this~\cite{DBLP:conf/aaai/YanXL18,2020Feedback,2019Symbiotic,2020Learning} have been generated. Liu et al.~\cite{2020Disentangling} proposed MS-G3D, which enables direct information propagation using intertemporal edges, making an effective improvement to spatio-temporal topological modelling. Chen et al.~\cite{2021Learning1} proposed a hop-aware hierarchical channel compression fusion layer based on the inspiration of feature compression~\cite{DBLP:conf/bmvc/ZouLWT20,2019AutoSlim,2021Hop}, which can effectively extract relevant information from neighboring nodes information from neighboring nodes efficiently. Dynamic skeletal topologies were also constructed to model the changes of actions over time better. Chen et al.~\cite{2021Channel} proposed a channel topology refinement graph convolution non-shared topology as well as the idea of dynamic graph convolution, which can spatially implement dynamic topology modeling.



{\bf Attention Mechanisms}. Methods with attention mechanisms are increasingly used in various fields, and HU et al.~\cite{2017Squeeze} proposed Squeeze-and-excitation networks, which are able to obtain scores on relevant dimensions. The transformer structure with self-attention~\cite{2017Attention} was proposed and applied to text-based tasks such as translation. With increasing attention, such methods started to be applied to the vision domain.

In skeletal-based action recognition tasks. Song et al.~\cite{2016An} first proposed spatial-temporal attention Long Short-
Term Memory method for modelling skeletal joint distinctions. Chiara et al.~\cite{2020sttr} proposed a novel Spatial-Temporal Transformer network (ST-TR), which uses spatio-temporal Transformer self-attention to model the representation of relationships between joints. Cheng et al.~\cite{2020Decoupling} added an attention guided drop mechanism to the model as a means of enhancing the regularisation effect and effectively improving the accuracy of action recognition. Shi et al.~\cite{2017Two} proposed an attention map method to represent the strength of the connection between two nodes. Ye et al.~\cite{2020Dynamic} proposed the Dynamic GCN method, which also proposed an attention mechanism method to capture spatial dependencies. Qiu et al.~\cite{2022Spatioxi} proposed the spatio-temporal tuples transformer method to capture dependencies between different joints. In the recent work, song et al~\cite{song2022constructing} proposed a Spatial Temporal
Joint Attention module that enables the finding of key joints for spatio-temporal sequence graphs to better achieve effective modelling of topology.

\begin{figure*}[t] 
\centering 
\includegraphics[width=0.8\linewidth,height=36\baselineskip]{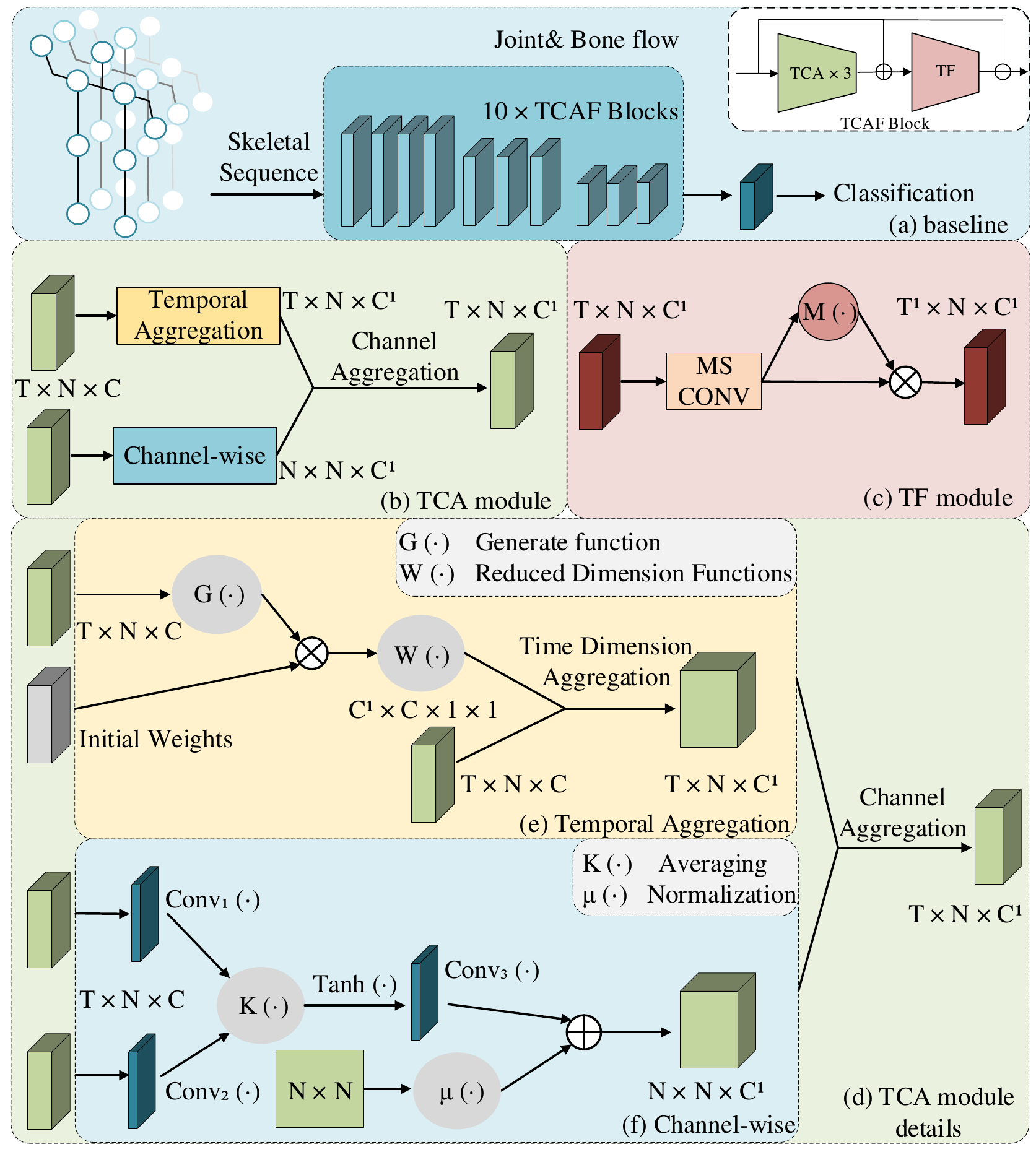} 
\caption{The pipeline of the proposed method. Our proposed network consists of 10 TCAF blocks, and each TCAF block consists of three TCA modules and one TF module. The TF module consists of multi-scale skeleton feature fusion containing MSCONV and feature fusion with attention mechanism. Detail structure of the TCA module: The TCA module consists of three parts. It contains temporal aggregation, channel-wise topology modeling and channel aggregation.} 
\label{11} 
\end{figure*}

\section{Method}
\label{sec:blind}

In this section, we introduce the proposed skeleton-based action recognation method as shown in Figure~\ref{11}(a). Section~\ref{sec:A} describes the relevant preparatory work, including the data representation and the graph convolution structure used. We detail the specific architecture of the used TCA module in Section~\ref{sec:B}. In Section~\ref{sec:C}, we provide a detailed description of the TF module. Finally, we illustrate our fusion method for different streams in Section~\ref{sec:E}. 

\subsection{Preliminaries}
\label{sec:A}
\noindent
{\bf Skeletal Data.} The graph on the skeletal sequence can be represented by $G=\left( V,E,X \right)$~\cite{DBLP:conf/aaai/YanXL18,2020sttr,2019Symbiotic}. The set $V$ represents all the nodes. The set of edges is denoted by $E$, the degree of connectivity between different joints is usually represented by the adjacency matrix $A\in R^{N\times N}$. The set of feature $X\in R^{N\times C}$ is represented as an input feature. Its dimension is the number of joints. In addition, we use a spatial partitioning method based on centrifugal and centripetal motions~\cite{DBLP:conf/aaai/YanXL18}, the spatio-temporal skeleton is spatially partitioned, which is represented as $(D, V, V)$, where $D$ is the number of delineated joint sets. \\
\noindent
{\bf Skeleton-based Graph Convolution.} Since the birth of graph convolution~\cite{2016Semi}, based on the ability to effectively model skeletal spatio-temporal maps, it has made a wide range of applications in the field of action recognition~\cite{DBLP:conf/aaai/YanXL18}:

\begin{align}
f_{out}(v_{i})=\sum _{v_{j}\in B_{i}}\frac{1}{Z_{ij}}f_{in}(v_{j})\cdot w(l_{i}(v_{j})),
\end{align}
where $f_{out}(v_{i})$ is output characteristics of the i-th joint, $ B_{i}$ is the  set of neighborhoods divided by the partitioning method, $Z_{ij}$ is equal to the number of bases in the corresponding subset, $w(l_{i}(v_{j}))$ is weighting function by index tensor. We used the spatial configuration partitioning method for skeletal sequence segmentation, $l_{i}(v_{j})$ can be divided into three different values~\cite{DBLP:conf/aaai/YanXL18,2020Feedback,2019Symbiotic,2020Learning,2021Channel}.

\subsection{TCA Module}
\label{sec:B}
\noindent
{\bf Channel-wise Topology Modeling.} In this part, we focus on spatial skeletal features. We use channel-wise topology modeling~\cite{2021Channel} to dynamically acquire the channel topology of a space, which mainly consists of correlation modeling function $\mathcal{M}(\cdot)$ and refinement function $\mathcal{R} (\cdot)$, where the correlation modeling function uses the input features $X \in R^{T \times N\times C}$ to model the channel correlations between joints to obtain channel-specific correlations $Q\in R^{N \times N\times C^{1}}$, and the refinement function sums the channel-specific correlations $Q$ with the adjacency matrix obtained from spatial configuration partitioning to obtain channel-wise topologies $S\in R^{N \times N\times C^{1}}$, the framework of which is shown in Figure~\ref{11}(f). In this paper, we use the following specific method formula~\cite{2021Channel}:

\begin{align}
S=\alpha \cdot Q+  \mu \left ( A_{k}  \right ),
\end{align}
where $A_{k}$ is k-th channel adjacency matrix, $\mu(\cdot)$ is expressed as the normalization of the third-order adjacency matrix and the dimensional transformation operation and $\alpha$ is a trainable parameter to represent the joint connection strength.


\noindent
{\bf Temporal Aggregation.} In this part, we mainly consider the time dimension skeletal feature processing, inspired by Temporally-Adaptive Convolutions~\cite{DBLP:journals/corr/abs-2110-06178}, we propose the temporal adaptive weight aggregation method, which is capable of generating calibrated weights in the temporal dimension based on the input features and aggregating them with the prior topology in time dimension to complete the temporal dynamic topological representation. Its framework is shown in Figure~\ref{11}(e). Specifically, we first generate the temporal weights by using the skeletal sequence as the input $X \in R^{T \times N\times C}$, using the generating function to multiply with the initial weights and perform the dimensional transformation to generate the temporal weights, which are temporally aggregated with the prior topology to obtain the new high-dimensional feature representation $A_{out} \in R^{T \times N\times C^{1}}$, as shown in the following equation:

\begin{align}
A_{out}=TA(W(W),X)=(W_{1}X_{1})\parallel...\parallel(W_{T}X_{T}),
\end{align}
where $W_{i}X_{i}$ is specified as:

\begin{align}
(W_{i}X_{i})_{ab}=\sum_{j}^{C} w(j)\cdot x(j),
\end{align}
where $TA(\cdot)$ indicates aggregation function, $\parallel$ indicates aggregation operation, $W(\cdot)$ is the dimensionality reduction function, $W(W)\in R^{C\times  C^{1}\times T}$ is the temporal weight feature, whose expression is~\cite{DBLP:journals/corr/abs-2110-06178}:

\begin{align}
W=\alpha _{t}\cdot W_{0},
\end{align}
where $\alpha _{t}=G(x_{1},...,x_{t})$~\cite{DBLP:journals/corr/abs-2110-06178}, and $\alpha _{t}$ is the calibrated weights with output channels, we use a different dimensional approach based on skeletal features, $W_{0}\in R^{  C^{1}\times C\times 1\times 1}$ is the initial weight.

\noindent
{\bf Channel Aggregation.} In this part, we focus on aggregating the spatio-temporal features of the first two parts to complete the balancing process, we aggregate them in the channel dimension, whose graph convolutional representation is formulated as shown below:

\begin{align}
f_{out}(v_{i})=\sum _{v_{j}\in B_{i}}A_{out}(v_{j})\cdot S(v_{j}),
\end{align}
where $A_{out}$ denotes the output feature of adaptive aggregation, $S$ denotes the output feature of channel-wise topology, $v_{i}$ denotes the i-th joint, the aggregated ground in the channel dimension is specified as follows:

\begin{align}
F_{out}=CA(A_{out},S)=(A_{out1}S_{1})\parallel...\parallel(A_{outC^{1}}S_{C^{1}}),
\end{align}
where $A_{outj}S_{j}$ is specified as:

\begin{align}
(A_{outj}S_{j})_{ab}=\sum_{i}^{N} a_{out}(i)\cdot s(i),
\end{align}
where $CA(\cdot)$ indicates channel aggregation function, $N$ denotes the number of joints, and $F_{out} \in R^{T \times N\times C^{1}}$ is the final output result.
\subsection{TF Module}
\label{sec:C}
\noindent
{\bf Multi-scale Skeleton Feature Fusion.} For feature fusion problems in temporal modeling, we use a method with an attention mechanism~\cite{2020Attentional} for feature fusion, unlike that, we use an input branch, which is used to solve the problem of feature fusion of contextual skeletal feature information to improve the effectiveness of modeling. For the temporal modeling feature, we use multi-scale convolutional~\cite{2021Channel} learning to augment the temporal convolutional layer. As shown in Figure~\ref{11}(c), the specific equation is expressed as:

\begin{align}
Z_{out} =sk\left ( MSCONV\left ( F_{out}  \right )   \right ), 
\end{align}
where $F_{out}$ denotes the features after TCA module, $MS CONV(\cdot)$~\cite{2021Channel} is the multi-scale convolution, $sk(\cdot)$ is an attentional feature fusion method with a single branch input (i.e. $MSCONV\left (\cdot  \right )\cdot M\left ( \cdot \right )$). The specific formula for $M\left ( \cdot \right )$~\cite{2020Attentional} is expressed as:

\begin{align}
 Sig(\mathbf{l}(\mathbf{Z}) \oplus \mathbf{g}(\mathbf{Z})),
\end{align}
where $Z$ is the input feature matrix, $g(\cdot)$ and $l(\cdot)$ is the global channel context and local channel context, respectively.


\begin{algorithm}[tb]
\caption{Dynamic fusion of models solver.}
\label{tab:al}
\textbf{Input}: four streams of action score: $\left \{ r_{lli} \right \} _{ll=1}^{4}$; $Label_{i}$: the true label of the action.\\
\textbf{Output}: Action accuracy.
\begin{algorithmic}[1] 
\STATE Initialize the fusion weights $\left ( a,b,c,d \right ) \longleftarrow $ the previous method;
\STATE Import the test scores of the four streaming frameworks $\left \{ r_{ki} \right \} _{k=1}^{4}$;
\WHILE{b$>$a$>$c$>$d and $\left ( a,b,c,d \right ) \longleftarrow$ (0,1]}
\STATE Calculate the correct rate of action prediction acc from Eq.~\ref{g1}, calculate r and the correct number of right according to the Eq.~\ref{g2};
\IF {the model accuracy improve}
\STATE Save new fusion weights and results;
\ELSE
\STATE Keep the parameters;
\ENDIF
\ENDWHILE
\STATE \textbf{return} acc;
\end{algorithmic}
\end{algorithm}


\subsection{Dynamic Fusion of Models}
\label{sec:E}

In order to obtain higher accuracy, many methods~\cite{2021Channel, 2020Skeletonshift, 2020Dynamic} nowadays use uniform fusion weights and use four streams of fusion: bone, bone motion, joint and joint motion. This leads to the problem of difficulty in reflecting the differences between different datasets and the inability to use all flow models. Therefore, to better address the variability between datasets as well as to better utilize all the flow models, we dynamically adjust the four flow models to achieve the better results. We established specific {\bf dynamic fusion planning equations}, the details of which are placed in the supplementary material, and eventually obtained accuracy by Algorithm~\ref{tab:al}.





\section{Experiments}
In this section, we perform an experimental evaluation of our model in skeleton-based action recognition. In Section~\ref{sec:AA}, We conducted experiments on three datasets namely Northwestern-UCLA~\cite{DBLP:journals/corr/WangNXWZ14}, NTU RGB+D~\cite{2016NTU}, and NTU RGB+D 120~\cite{2020NTU}. 
Section~\ref{sec:AB} describes the specific configuration used to conduct our experiments. In Section~\ref{sec:CA}, to validate the generality of our model, we perform a final evaluation on these three datasets and compare them with other existing state-of-the-art methods. In Section~\ref{sec:DA}, We provide ablation experiments to discuss the impact of the different components of our proposed approach, the ablation experiments are placed on the supplementary material. 




\subsection{Datasets}
\label{sec:AA}
\noindent
{\bf Northwestern-UCLA.} Northwestern-UCLA
dataset~\cite{DBLP:journals/corr/WangNXWZ14} is captured from multiple angles with three cameras. The dataset contains 1,494 video clips containing 10
action categories, which are performed by ten different objects. The training set used was from two of the cameras and the test set was from the other Kinect camera.

\noindent
{\bf NTU RGB+D}. NTU RGB+D~\cite{2016NTU}  contains 60 categories of actions, with a total of 56,880 samples, of which 40 categories are daily behavioral actions, 9 categories are health-related actions, and 11 categories are two-player mutual actions. 60 categories of actions are performed by 40 people ranging in age from 10 to 35 years old. It was acquired by the Microsoft Kinect v2 sensor and used three different camera angles, and in this paper we use data in the form of 3D skeletal information. It uses two criteria in dividing the training and test sets: (1) cross-subject (X-sub): the specified 20 people are used as the training set and the rest as the test set. (2) cross-view (X-view): the data collected by camera 1 is used as the training set, and the data collected by the remaining two cameras are used as the test set.

\noindent
{\bf NTU RGB+D 120.} NTU RGB+D 120~\cite{2020NTU} is the largest 3D joint dataset of human movements available. This dataset expands the NTU RGB+D dataset by adding 57,367 skeletal sequences and 60 additional action categories. This dataset was captured with three cameras, with a total of 113,945 samples, by 106 volunteers. The dataset was also divided between the training and test sets using two different division criteria: (1) cross-subject (X-sub): the training data is from 53 subjects, and the test set is from the remaining subjects. (2) cross-setup (X-setup): the training data set from samples with even set IDs, and the test data set are from samples with other set IDs.

\begin{table}[h]
 \centering
\begin{tabular}{lllll}
\hline\noalign{\smallskip}
Methods & N-UCLA
 \\
\noalign{\smallskip}\hline\noalign{\smallskip}
Lie Group~\citeyearpar{VivekVeeriah2015DifferentialRN}&74.2\\
HBRNN-L~\citeyearpar{2015Hierarchical}&78.5\\
Glimpse Clouds~\citeyearpar{2018Glimpse}&87.6\\
VA-fusion~\citeyearpar{2018View}&88.1\\
Action Machine~\citeyearpar{2018Action}&92.3\\
AGC-LSTM~\citeyearpar{2019An}&93.3\\
SGN~cite~\citeyearpar{2020Semantics}&92.5\\
Shift-GCN~\citeyearpar{2020Skeletonshift}&94.6\\
DC-GCN+ADG~\citeyearpar{2020Decoupling}&95.3\\
CTR-GCN~\citeyearpar{2021Channel}&96.5\\
Ta-CNN~\citeyearpar{xu2022topology}&96.1\\

Ta-CNN+~\citeyearpar{xu2022topology}&{\bf97.2}\\ 
\noalign{\smallskip}\hline
TCA-GCN& 96.8\\
TCA-GCN(4sD)& \underline{97.0}\\

\noalign{\smallskip}\hline
\end{tabular}
\caption{Our model is compared with the state-of-the-art approach on the NW-UCLA dataset, where 4sD denotes the result of four-stream fusion in Section~\ref{sec:E}.}
\label{tab:31}
\end{table}

\begin{table}[ht]
\centering
\begin{tabular}{lllll}
\hline\noalign{\smallskip}
Methods & X-Sub &X-View\\
\noalign{\smallskip}\hline\noalign{\smallskip}
ST-LSTM~\citeyearpar{2016Spatio}&69.2&77.7\\
ST-GCN~\citeyearpar{DBLP:conf/aaai/YanXL18}&81.5&88.3\\
RA-GCNv1~\citeyearpar{2019Richly}&85.9&93.5\\
AS-GCN~\citeyearpar{2019Actional}&86.8&94.2\\
2s-AGCN~\citeyearpar{2018Two}&88.5&95.1\\

Shift-GCN~\citeyearpar{2020Skeletonshift}&90.7&96.5\\
DDGCN~\citeyearpar{20201Skeleton} & 91.1 & {\bf 97.1} \\
Dynamic GCN~\citeyearpar{2020Dynamic}&91.5&96.0\\
MST-G3D~\citeyearpar{2020Disentangling}& 91.5 & 96.2 \\
MST-GCN~\citeyearpar{2021Multi} & 91.5 & 96.6 \\
Skeletal-GNN~\citeyearpar{2021Learning}&91.6&96.7\\
4s DualHead-Net~\citeyearpar{2021Learning1}&92.0&96.6\\
CTR-GCN~\citeyearpar{2021Channel}&92.4&96.8\\
Ta-CNN~\citeyearpar{xu2022topology}&90.4&94.8\\
Ta-CNN+~\citeyearpar{xu2022topology}&90.7&95.1\\
EfficientGCN-B4~\citeyearpar{2021Constructing}&91.7&95.7\\
\noalign{\smallskip}\hline
TCA-GCN&\underline{92.6}&96.9\\
TCA-GCN(4sD)&{\bf 92.8}&\underline{97.0}\\
\noalign{\smallskip}\hline
\end{tabular}
\caption{Our model is compared with the state-of-the-art method on the NTU RGB+D dataset, where 4sD denotes the result of four-stream fusion in Section~\ref{sec:E}.}
\label{tab:32} 
\end{table}

\begin{table}[ht]
 \centering
\begin{tabular}{lllll}
\hline\noalign{\smallskip}
Methods &X-Sub&X-Set\\
\noalign{\smallskip}\hline\noalign{\smallskip}
ST-LSTM~\citeyearpar{2016Spatio}&55.7&57.9\\
FSNet~\citeyearpar{2019Skeleton12}&59.9&62.4\\
RA-GCNv1~\citeyearpar{2019Richly}&74.4&79.4\\
2s-AGCN~\citeyearpar{2018Two}&82.9&84.9\\
Shift-GCN~\citeyearpar{2020Skeletonshift}&85.9&87.6\\
MST-G3D\citeyearpar{2020Disentangling} &86.9&88.4 \\
Dynamic GCN~\citeyearpar{2020Dynamic}&87.3&88.6\\
MST-GCN~\citeyearpar{2021Multi} &87.5&88.8 \\
Skeletal-GNN~\citeyearpar{2021Learning}&87.5&89.2\\
4s DualHead-Net~\citeyearpar{2021Learning1}&88.2&89.3\\
CTR-GCN~\citeyearpar{2021Channel}&88.9&90.6\\
Ta-CNN~\citeyearpar{xu2022topology}&85.4&86.8\\
Ta-CNN+~\citeyearpar{xu2022topology}&85.7&87.3\\
EfficientGCN-B4~\citeyearpar{2021Constructing}&88.3&89.1\\
\noalign{\smallskip}\hline
TCA-GCN&\underline{89.2}& \underline{90.7}\\
TCA-GCN(4sD)&{\bf 89.4}&{\bf 90.8}\\
\noalign{\smallskip}\hline
\end{tabular}
\caption{Our model is compared with the state-of-the-art method on the NTU RGB D+120 dataset, where 4sD denotes the result of four-stream fusion in Section~\ref{sec:E}.}
\label{tab:33}
\end{table}

\subsection{Implementation Details}
\label{sec:AB}
All experiments are conducted  with the PyTorch deep learning framework, and we use three Tesla M40 GPUs. Cross-entropy method is used as the loss function. Our models are trained with Stochastic Gradient Descent (SGD) with momentum(0.9).  When training the model on the three datasets, we used the warmup strategy for the first 5 epochs. About learning rate, we set to 0.1 and use decays at 35 epoch and 55 epoch, ending the training at 65 epoch. For the NTU-RGB+D 60$\&$120 dataset, we used the preprocessing~\cite{2020Semantics}, setting the batch size to 64. For the Northwestern-UCLA dataset, we set the batch size to 16. In multi-stream fusion for different datasets, applying the previous fusion method~\cite{2021Channel, 2020Skeletonshift, 2020Dynamic}.

\subsection{Comparison with the State-of-the-Art}
\label{sec:CA}
In order to fairly compare the results, we not only used the previous multi-stream fusion model~\cite{2021Channel,DBLP:conf/cvpr/Cheng0HC0L20,2020Dynamic}. In addition, we have also included the results of Section~\ref{sec:E} experiments named 4sD. The details of the implementation are included in the supplementary material. 

\noindent
{\bf Comparison with SOTAs.} As shown in Table~\ref{tab:31},~\ref{tab:32},~\ref{tab:33}, our models obtain state-of-the-art results on almost all benchmarks. In NTU60, our method takes into account the dynamic features in time to better demonstrate the spatio-temporal variation of skeletal sequences and to better achieve the degree of differentiation of features.
Table~\ref{tab:32} shows our accuracy is 0.2$\%$ and 0.1$\%$ higher than CTR-GCN~\cite{2021Channel} in X-Sub and X-View criteria, respectively.
In NTU120, our method is not only able to consider spatio-temporal features and effectively combine time-space modeling, but also has more advantages for large dataset identification. Table~\ref{tab:33} shows our accuracy is 0.3$\%$ and 0.1$\%$ higher than CTR-GCN~\cite{2021Channel} in X-Sub and X-Set criteria, respectively.

Finally, after conducting the new fusion experiments, our model: TCA-GCN(4sD), it has improved to some extent over all previous fusion methods. For example, our accuracy is 0.5$\%$ higher than state-of-the-art method in NTU120. This is because we not only make full use of the stream framework, but also can perform dynamic weight fusion according to the importance of different streams.


\begin{figure}[htp]
    \includegraphics[width=8cm,height = 6cm]{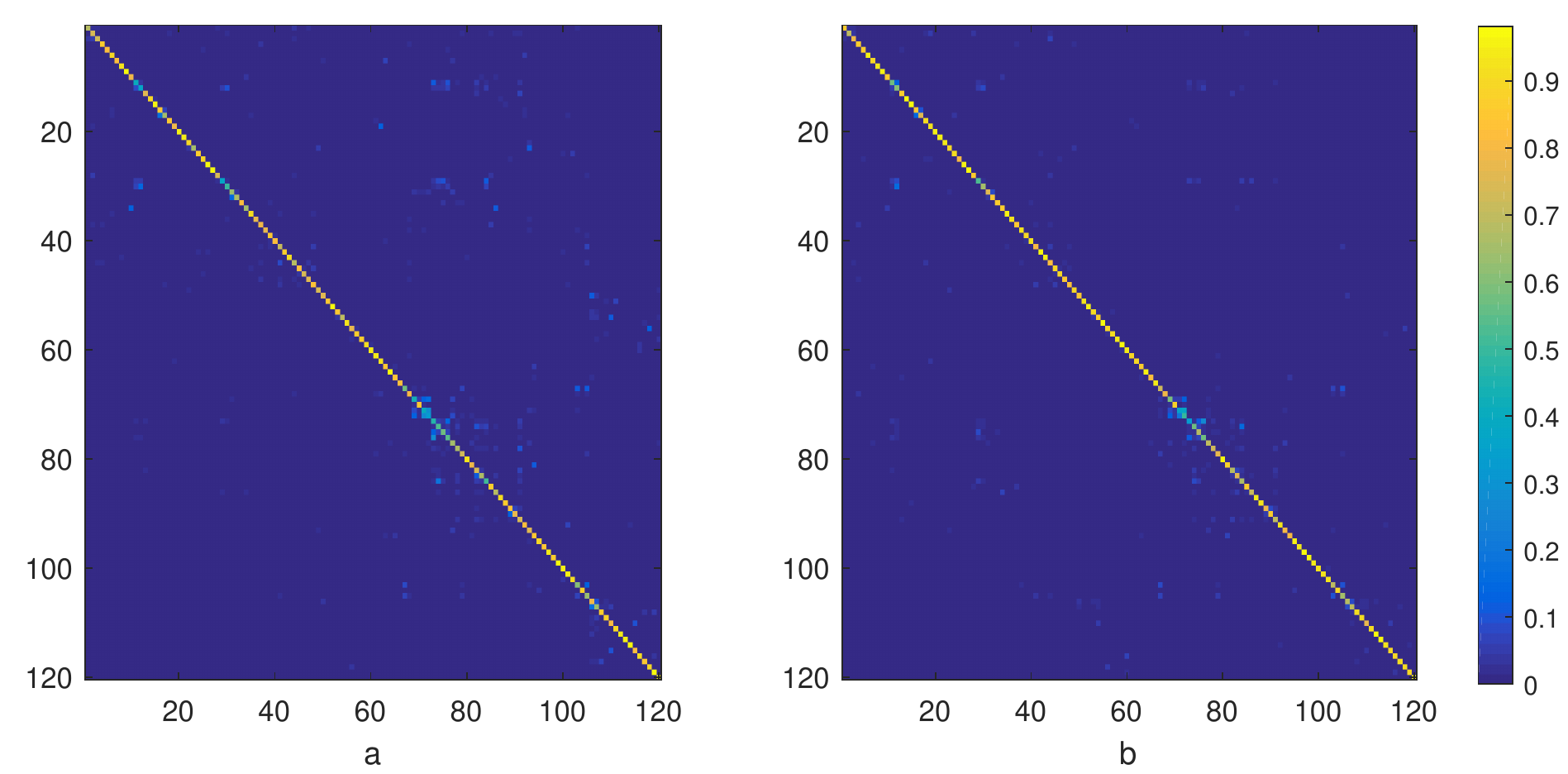}
    \caption{Confusion matrix with attention mechanism and our methods. (a), (b) are S-TR and our method (only joint).}
    \label{fig:galaxyat}
\end{figure}
\begin{table}[htp]
 \centering
\begin{tabular}{llllll}
\hline\noalign{\smallskip}
Action&T-TR&S-TR&Our(J)&E-B0&Our\\
\noalign{\smallskip}\hline\noalign{\smallskip}
eat meal&0.62&0.75&0.71&0.76&{\bf 0.86}\\
clapping&0.78&0.79&{\bf0.89}&0.80&{\bf0.89}\\
put on a shoe&0.75&0.85&{\bf0.87}&0.78&0.85\\
flick hair&0.80&0.80&0.80&0.86&{\bf0.96}\\
make OK &0.44&0.40&0.37&0.69&{\bf0.73}\\
make victory &0.38&0.34&0.51&0.57&{\bf0.63}\\
fold paper&0.77&0.74&0.69&0.65&{\bf0.84}\\
put into bag&0.84&0.83&0.81&0.80&{\bf0.91}\\
hit with object&0.76&0.79&0.69&0.71&{\bf0.86}\\
hoot with gun&0.81&0.81&0.79&0.77&{\bf0.92}\\
\noalign{\smallskip}\hline
\end{tabular}
\caption{Accuracy of specific action classification under different methods, where E-B0 represents for EfficientGCN-B0 and put into bag represents for put object into bag.}
\label{tab:12at}
\end{table}


%

\noindent
{\bf Comparison of attentional mechanisms.} To explore the advantages of our model over methods with attentional mechanisms, the model is compared with other methods with attentional mechanisms. We perform experimental comparisons not only on single streams, but also on multi-stream fusion in terms of specific actions. The experimental results show that, as shown in Figure~\ref{fig:galaxyat}: In terms of accuracy for single streams, our model fares better than S-TR~\cite{2020sttr} in terms of full action recognition, which is particularly evident in the top right corner of the figure, indicating that our method also has some advantages in terms of some similar action recognition, and that our method is better able to learn topologically on single streams. This suggests that our spatio-temporal aggregation model has a better advantage, mainly because our model is able to model the spatio-temporal topology effectively and better learn effectively based on the topology. The results of specific actions are shown in Table~\ref{tab:12at}: (1) In general, our model has some advantages in recognizing difficult actions, for example, our model is 2.4$\%$, 1.1$\%$, and 1.0$\%$ higher than other methods in eating meal, which is mainly attributed to our spatio-temporal topology modeling can better solve the effective extraction of spatio-temporal features. (2) From the individual approaches, our method is not only able to surpass the attentional methods in time and space in a single stream, but also has certain advantages in the combination of multiple streams. Finally, the experimental accuracy is further improved by comparing both approaches, which indicates that our temporal aggregation approach not only integrates well with the attention mechanism, but also shows better performance than the attention mechanism. This is because for skeletons, aggregation is better at modeling the features. 
\subsection{Ablation Study}
\label{sec:DA}

We demonstrate the advantages of our proposed model and the effectiveness of each component. Since the NTU RGB+D 120 data is the largest 3D joint annotation dataset for human motion recognition and is more general and researchable, we describe this data as two benchmarks unless otherwise stated. We put the experiment in the {\bf supplementary material}.

\section{Conclusion}
In this paper, we propose a temporal-channel dimension aggregation-based graph neural network that includes TCA module and TF module. The network is capable of dynamic spatio-temporal topological feature extraction and their effective aggregation, as well as attention feature fusion for multi-scale features, to complete the modeling more effectively. Extensive experiments show that our model has advanced performance on different types of datasets.

\bibliography{aaai23}
\clearpage
\appendix 
\section{Supplementary Material}
\subsection{Specific Configuration of Model Architecture}
{\bf Modeling of Spatial Dimensions.} In the spatial modeling module, we use three TCA modules for dynamic spatio-temporal feature extraction, and we use the output feature results of this module as the input features  X $\in R^{T \times N\times C}$ of the temporal modeling module. Specifically, channel-wise topology modeling is used for spatial skeletal topology optimization modeling to realize the combination of spatial a priori skeletal knowledge and a posteriori knowledge, and the temporal aggregation method is used to complete the high-dimensional feature representation with temporal adaptive aggregation, and then they are channel aggregated, and finally, the results of the three channels of this module are summed to form the final output features. 

\noindent
{\bf Modeling of Temporal Dimensions.} We use the multi-scale convolutional skeleton feature fusion method as a temporal modeling module (TF module) that can consider action semantics for different skeletal features, better feature fusion of skeletal information, and improved modeling capabilities. The module takes $F_{out} \in R^{T \times N\times C^{1}}$ as input and obtains the output $Z_{out} \in R^{T \times N\times C^{1}}$, the framework of which is shown in Figure~\ref{11}(a). Finally, the spatial modeling module is connected with the temporal modeling module for residuals to form the TCAF block. Our network uses ten TCAF blocks.

\subsection{Dynamic Fusion Planning Equations}
The data flow fusion problem can be viewed as a functional optimization process, where the optimal fusion effect is the objective function and the individual flow models are used as decision variables to construct the fusion-based nonlinear programming equations. Here we use the control variable method. We carried out a quantitative process for making full use of the four streams fusion weights for comparison, we put the initial values of the four weights greater than 0 into the model constraint part to achieving the importance of different streams.  We set four relations with small weights and put them into constraints. In addition, we output the maximum accuracy that satisfies the condition, the pseudo-nonlinear programming equation is shown below:

\begin{align}
\label{g1}
goal: max(right/zong),
\end{align}
\begin{align}
\label{g2}
\begin{cases}
 & \text y_{i}=a\cdot r_{11i}+b\cdot r_{22i}+c\cdot r_{33i}+d\cdot r_{44i}, \\ 
 & \text right=Sum\left ( Max\left ( Index\left ( y_{i}\right )\right )==Label_{i}\right ), \\ 
 & \text r_{11i}\leftarrow r_{1i},r_{22i}\leftarrow r_{2i},r_{33i}\leftarrow r_{3i},r_{44i}\leftarrow r_{4i},\\
 & \text b>a>c>d, \\
 & \text a,b,c,d\in \left ( 0,0.05,1\right ],\\
 & \text i=1,2,...,zong,
\end{cases}
\end{align}
where $y_{i}$ denotes the multi-stream fusion score, $a,b,c,d$ denotes the weights of different streams respectively, $r_{11i}$ denotes the score of a single stream, $Label_{i}$ denotes the true label of the action, $Index(\cdot)$ is the index value where the solved action score is located, in order to be able to correspond with the true value. $\leftarrow$ denotes the extraction operation operator, $zong$ denotes the number of actions for the test set, $right$ indicates the number of correct predicted actions. To facilitate efficient solving of nonlinear programming equations, we set up a pseudo code, as shown in the Algorithm~\ref{tab:al}.
\subsection{Efficient of Multi-scale Skeleton Feature Fusion}
 To verify the efficient of multi-scale convolutional skeleton feature fusion in NTU60(X-Sub)~\cite{2016NTU}, we compared its time consumption with the current state-of-the-art model~\cite{2021Channel}. As shown in the Table~\ref{tab:111}, The experimental results show that multi-scale skeleton feature fusion is able to reduce the number of blocks in the model, resulting in a lower perceptual field, and in terms of time consumption, multi-scale skeleton feature fusion is able to accelerate the convergence speed. In practice, this method can significantly reduce the application time and increase the efficiency, which also shows efficiency of our proposed method.
 
\begin{table}[htp]
 \centering
\begin{tabular}{llccc}
\hline\noalign{\smallskip}
Model & CTR-GCN &TF module with CTR-GCN\\
\noalign{\smallskip}\hline\noalign{\smallskip}
Block&10&{\bf7}\\
Time(epoch)& 0&{\bf -2(min)}\\
Time(all)& 0&{\bf -2(hours)}\\
Acc&89.6&89.6\\

\noalign{\smallskip}\hline
\end{tabular}
\caption{Comparison of efficiency with state-of-the-art models. Experiments on the same GPU.}
\label{tab:111}
\end{table}

\subsection{Two Interactive Performance}
\begin{figure}[t]
\centering  
{
\label{Fig.sub.1}
\includegraphics[width=3cm,height = 3cm]{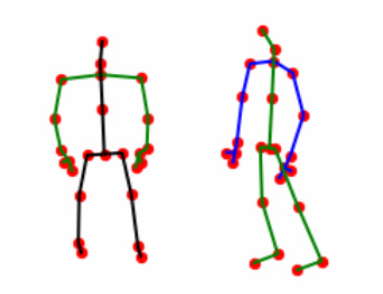}}
{
\label{Fig.sub.2}
\includegraphics[width=3cm,height = 3cm]{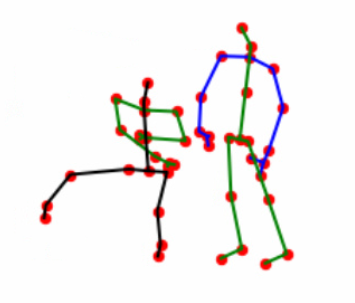}}
{
\label{Fig.sub.3}
\includegraphics[width=3cm,height = 3cm]{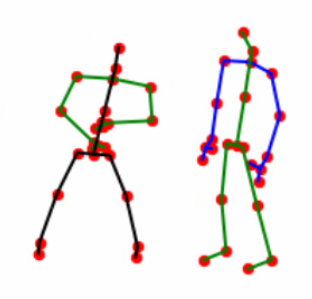}}
\caption{The skeleton visualization of knock over action. For the sake of conciseness and clarity of the general change process, three time periods are selected for display in accordance with the chronological order of the action. The blue color indicates the upper limb part of the skeletal frame. The person on the left has a larger change in movement, while the other person on the right has a smaller.}
\label{1121}
\end{figure}

For some skeletal sequences, there are two people moving with each other. The movements have occlusion as well as the difference in spatio-temporal position, which causes the problem of multi-semantic movements and makes it difficult to recognize such movements. For example, the action in Figure~\ref{1121}, it can be found that one person's action changes more obviously, and the other person's action is not obvious. To verify how well our model handles this problem, we conducted experiments on bone streams and compared the results with the current best method~\cite{2021Channel}. Our model improves the effect of such actions as knock over and follow by 5$\%$ and 4$\%$. This is because our proposed method is able to model multi-scale information with a priori skeletal knowledge to solve the action semantics problem. This implies the feasibility of our method to deal with this type of problem.

\begin{table*}[ht]
 \centering
\begin{tabular}{cccclll}
\hline\noalign{\smallskip}
Data Benchmark                                     & \multicolumn{2}{c}{Number of Streams}             &\multicolumn{2}{c}{Same data Set Configuration}  & \multicolumn{2}{c}{Using Dynamic Fusion}                \\ \hline
                                                   & 3                      & 4                      & Same                   & \multicolumn{1}{c}{Not} & \multicolumn{1}{c}{Not}   & \multicolumn{1}{c}{Dynamic} \\ \hline
\multicolumn{1}{c}{} & \multicolumn{1}{c}{$\surd$} & \multicolumn{1}{c}{}  & \multicolumn{1}{c}{$\surd$} & \multicolumn{1}{l}{}   & \multicolumn{1}{l}{92.4$\star 
$} & \multicolumn{1}{l}{}       \\ 
\multicolumn{1}{c}{{NTU60(X-Sub)}}                              & \multicolumn{1}{c}{$\surd$} & \multicolumn{1}{c}{}  & \multicolumn{1}{c}{$\surd$} & \multicolumn{1}{l}{}   & \multicolumn{1}{l}{92.6} & \multicolumn{1}{l}{}       \\ 
\multicolumn{1}{c}{}                              & \multicolumn{1}{l}{}  & \multicolumn{1}{c}{$\surd$} & \multicolumn{1}{c}{}  & \multicolumn{1}{c}{$\surd$}  & \multicolumn{1}{l}{}     & \multicolumn{1}{c}{92.8}   \\ \hline

\multicolumn{1}{c}{} & \multicolumn{1}{c}{$\surd$} & \multicolumn{1}{c}{}  & \multicolumn{1}{c}{$\surd$} & \multicolumn{1}{l}{}   & \multicolumn{1}{l}{96.8$\star 
$} & \multicolumn{1}{l}{}       \\ 
\multicolumn{1}{c}{{NTU60(X-View)}}                              & \multicolumn{1}{c}{$\surd$} & \multicolumn{1}{c}{}  & \multicolumn{1}{c}{$\surd$} & \multicolumn{1}{l}{}   & \multicolumn{1}{l}{96.9} & \multicolumn{1}{l}{}       \\ 
\multicolumn{1}{c}{}                              & \multicolumn{1}{l}{}  & \multicolumn{1}{c}{$\surd$} & \multicolumn{1}{c}{}  & \multicolumn{1}{c}{$\surd$}  & \multicolumn{1}{l}{}     & \multicolumn{1}{c}{97.0}   \\ \hline

\multicolumn{1}{c}{} & \multicolumn{1}{c}{$\surd$} & \multicolumn{1}{c}{}  & \multicolumn{1}{c}{$\surd$} & \multicolumn{1}{l}{}   & \multicolumn{1}{l}{88.9$\star 
$} & \multicolumn{1}{l}{}       \\ 
\multicolumn{1}{c}{{NTU120(X-Sub)}}                              & \multicolumn{1}{c}{$\surd$} & \multicolumn{1}{c}{}  & \multicolumn{1}{c}{$\surd$} & \multicolumn{1}{l}{}   & \multicolumn{1}{l}{89.2} & \multicolumn{1}{l}{}       \\ 
\multicolumn{1}{c}{}                              & \multicolumn{1}{l}{}  & \multicolumn{1}{c}{$\surd$} & \multicolumn{1}{c}{}  & \multicolumn{1}{c}{$\surd$}  & \multicolumn{1}{l}{}     & \multicolumn{1}{c}{89.4}   \\ \hline

\multicolumn{1}{c}{} & \multicolumn{1}{c}{$\surd$} & \multicolumn{1}{c}{}  & \multicolumn{1}{c}{$\surd$} & \multicolumn{1}{l}{}   & \multicolumn{1}{l}{90.6$\star 
$} & \multicolumn{1}{l}{}       \\ 
\multicolumn{1}{c}{{NTU120(X-Set)}}                              & \multicolumn{1}{c}{$\surd$} & \multicolumn{1}{c}{}  & \multicolumn{1}{c}{$\surd$} & \multicolumn{1}{l}{}   & \multicolumn{1}{l}{90.7} & \multicolumn{1}{l}{}       \\ 
\multicolumn{1}{c}{}                              & \multicolumn{1}{l}{}  & \multicolumn{1}{c}{$\surd$} & \multicolumn{1}{c}{}  & \multicolumn{1}{c}{$\surd$}  & \multicolumn{1}{l}{}     & \multicolumn{1}{c}{90.8}   \\ \hline

\multicolumn{1}{c}{} & \multicolumn{1}{c}{} & \multicolumn{1}{c}{$\surd$}  & \multicolumn{1}{c}{$\surd$} & \multicolumn{1}{l}{}   & \multicolumn{1}{l}{96.5$\star 
$} & \multicolumn{1}{l}{}       \\ 
\multicolumn{1}{c}{{N-UCLA}}                              & \multicolumn{1}{c}{} & \multicolumn{1}{c}{$\surd$}  & \multicolumn{1}{c}{$\surd$} & \multicolumn{1}{l}{}   & \multicolumn{1}{l}{96.8} & \multicolumn{1}{l}{}       \\ 
\multicolumn{1}{c}{}                              & \multicolumn{1}{l}{}  & \multicolumn{1}{c}{$\surd$} & \multicolumn{1}{c}{}  & \multicolumn{1}{c}{$\surd$}  & \multicolumn{1}{l}{}     & \multicolumn{1}{c}{97.0}   \\ \hline

\end{tabular}
\caption{Comparison of dynamic nonlinear fusion settings. Where $\star$ denotes the SOTA model selected to be compared, Number of Streams indicates the number of stream used, Same data set configuration indicates whether to use the previous model fusion weights, Using Dynamic Fusion indicates whether weights are dynamically assigned for different streams.}
\label{tab:25}
\end{table*}
\subsection{Dynamic Fusion of Models}
To solve the problem of uneven use of the flow framework and the weight assignment problem, we use a mathematical approach to solve the problem, using CTR-GCN~\cite{2021Channel} and our proposed model with the same configuration as a control experiment. As can be seen from Table~\ref{tab:25}, our proposed dynamic fusion uses all four streams, and in addition, the weights we use are consistent with the real situation of each stream. For example, if the result of our bone mode is better than that of the joint mode, then it should be given more weight than the joint mode when fusion is performed, and the results of the method are improved on different data sets, which also indicates the good applicability of our method.
\subsection{Single-stream Comparative Ablation Experiments}
To compare the combined effect of our model based on four streams, including bone, joint, bone motion, and joint motion. Based on the NTU RGB+D 120 dataset, the comparison is made one by one with those of the publicly available state-of-the-art methods~\cite{2021Channel}, as shown in Table~\ref{tab:13}, our model is higher than the state-of-the-art methods by 1.1$\%$, 0.2$\%$, 0.5$\%$, 0.2$\%$ for the four streams, respectively. The efficiency as well as the applicability of our proposed skeleton-based action recognition model for each stream is verified.

\begin{table}[htp]
 \centering

\begin{tabular}{lllll}
\hline\noalign{\smallskip}
Stream & CTR-GCN &Ours\\
\noalign{\smallskip}\hline\noalign{\smallskip}
Bone&85.7&{\bf86.8}\\
Joint& 84.9&{\bf 85.1}\\
Bone motion& 81.2&{\bf 81.7}\\
Joint motion&81.4&{\bf81.6}\\

\noalign{\smallskip}\hline
\end{tabular}
\caption{A comparison of the combined performance of our models on different stream.}
\label{tab:13}
\end{table}

\subsection{Exploring Model Configurations}
To explore our model in dealing with the spatio-temporal sequence characteristics of the skeleton, we conducted experiments on the settings of the components, and the results are tabulated below, including the reduction rate r of the temporally channel adaptive aggregation, the activation function of each spatial modeling module, and the number of model blocks l. As shown in the Table~\ref{133}, after the experiments, we found that the results were better than the state-of-the-art methods~\cite{2021Channel}, which also confirmed the robustness of our method, as follows, (1) by modifying the activation function of the spatial modeling part, the results are very different and have good performance, indicating the applicability and compatibility of our method. (2) By modifying the activation function in the spatial modeling part, the difference in results is very small, and both have good performance, so we use this activation function considering that the ReLU function is easier to learn to optimize. (3) In addition, we also modeled the number of basic blocks because the sensory field of the joint is crucial for graph convolution, and the results show that having a better modeled sensory field at k=10, the larger the sensory field contains more advanced skeletal features, which is very detrimental to the recognition of subtle movements. We finally chose method A as our basic configuration, which has an accuracy of 86.8$\%$.

\begin{table}[htp]
 \centering
\begin{tabular}{lllll}
\hline\noalign{\smallskip}
Methods & r &k & $\sigma $ & Acc\\
\noalign{\smallskip}\hline\noalign{\smallskip}
CTR-GCN&$/$&10&$/$&85.7\\
A& 2&10&ReLU&86.8\\
B& 2&10&Tanh&86.8\\
C& 1&10&ReLU&86.4\\
D& 3&10&ReLU&86.6\\
E& 2&8&ReLU&86.2\\
F& 2&12&ReLU&86.3\\
\noalign{\smallskip}\hline
\end{tabular}
\caption{Comparison of the results of our model with different part settings.}
\label{133}
\end{table}

\begin{figure}[htp]
    \centering
    \includegraphics[width=8cm]{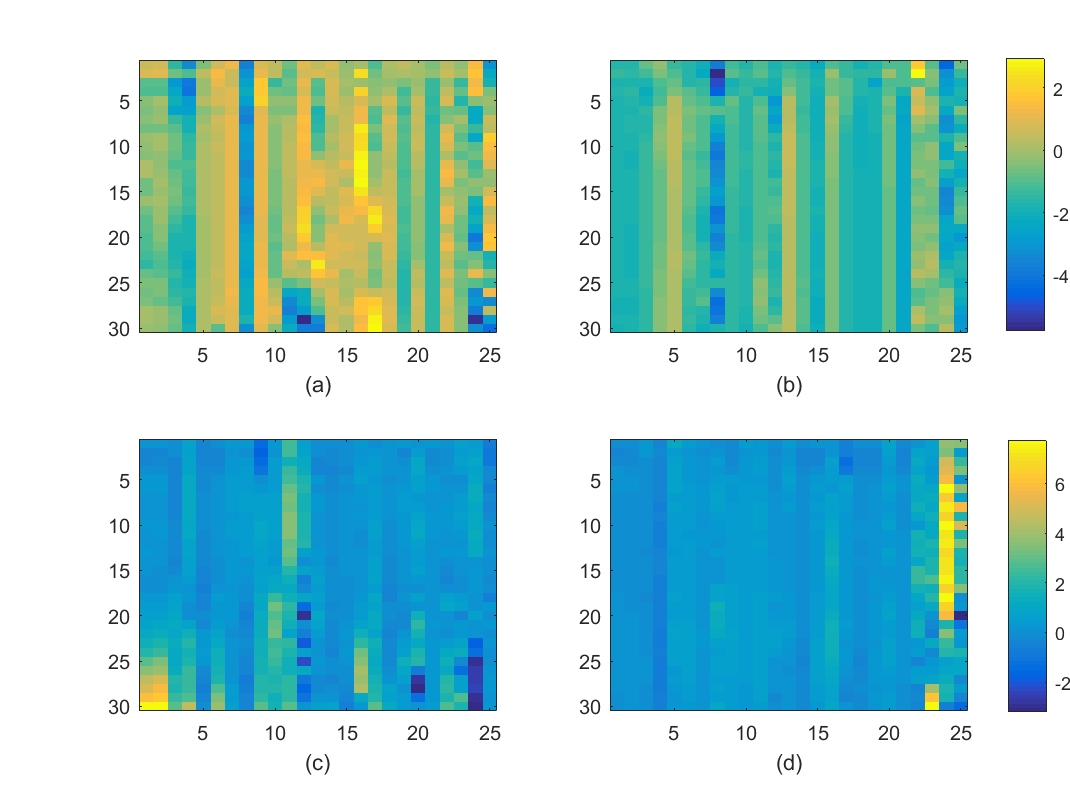}
    \caption{Visualization of joint features. (a), (b) are the previous feature transformation methods and (c), (d) is our proposed adaptive aggregation transformation method for skeletal sequences. The left side is squat down, the right side is eat meal, and we choose the same time dimension for visualization in order to show the convenience.}
    \label{fig:galaxy66}
\end{figure}

\subsection{Visualization of Joint Characteristics}
We show the feature-transformed nodal features of the two actions and the aggregated nodal features on the skeletal sequence proposed in this paper on the dataset, using the number of nodal points and the time series dimensions for feature visualization, respectively. The visualization results which are shown in Figure~\ref{fig:galaxy66}, where (a), (b) show the previously used method~\cite{2021Channel}, and (c), (d) show the method we used, and (b), (d) are the eating actions. We observe that (1) in terms of joint feature representation, our method has a wider range of features because our temporally channel adaptive aggregation method can dynamically adjust the weights and thus perform temporal aggregation, indicating that our method is better at representing joint features, indicating that our method is better at representing joints, (2) the visualization results we show are different for different actions, indicating that our method is able to perform adaptive aggregation learning for different actions, and (3) when the same action is represented, our model visualization is more differentiated, indicating that that our method is able to make dynamic adjustments to action timeliness based on time series. (4) From a practical point of view, the importance of the movement of the hands is obviously more expressive when performing eat meal, and when performing squat down, the characteristics of each part change significantly with time, which is also in line with the actual situation.






\end{document}